\documentclass[runningheads]{llncs}

\usepackage{graphicx}

\usepackage[super]{nth}
\usepackage{algorithm}
\usepackage{algpseudocode}
\usepackage{pgfplots}
\usepackage[final]{changes}
\usepackage{booktabs}
\usepackage{todonotes}
\PassOptionsToPackage{hyphens}{url}
\usepackage{hyperref}
\usepackage[title]{appendix}
\usepackage{fourier} 
\usepackage{array}
\usepackage{makecell}
\usepackage[utf8]{inputenc}

\pgfplotsset{compat=1.14}
\begin{document}

\algnewcommand\algorithmicswitch{\textbf{switch}}
\algnewcommand\algorithmiccase{\textbf{case}}
\algnewcommand\algorithmicassert{\texttt{assert}}
\algnewcommand\Assert[1]{\State \algorithmicassert(#1)}%
\algdef{SE}[SWITCH]{Switch}{EndSwitch}[1]{\algorithmicswitch\ #1\ \algorithmicdo}   {\algorithmicend\ \algorithmicswitch}%
\algdef{SE}[CASE]{Case}{EndCase}[1]{\algorithmiccase\ #1}{\algorithmicend\     \algorithmiccase}%
\algtext*{EndSwitch}%
\algtext*{EndCase}%

\title{From Textual Information Sources to Linked Data in the Agatha Project}
%
%

\author{Paulo Quaresma\orcidID{0000-0002-5086-059X} \and
Vitor Beires Nogueira\orcidID{0000-0002-0793-0003} \and
Kashyap Raiyani\orcidID{0000-0002-6166-2038} \and 
Roy Bayot\orcidID{0000-0002-1290-0239} \and 
Teresa Gonçalves\orcidID{0000-0002-1323-0249}}
\authorrunning{P. Quaresma et al.}
%
\institute{Universidade de \'Evora, Portugal \\
LISP - Laboratory of Informatics, Systems and Parallelism\\
\email{\{pq,vbn,kshyp,rkbayot,tcg\}@uevora.pt}}
\maketitle              
\begin{abstract}
Automatic reasoning about textual information is a challenging task in modern Natural Language Processing (NLP) systems.  In this work we describe our proposal for representing and reasoning about Portuguese documents by means of Linked Data like ontologies and thesauri. Our approach resorts to a specialized pipeline of natural language processing (part-of-speech tagger, named entity recognition, semantic role labeling) to populate an ontology for the domain of criminal investigations.
The provided architecture and ontology are language independent. Although some of the NLP modules are language dependent, they can be built using adequate AI methodologies.  

\end{abstract}

\section{Introduction}
\label{sec:introduction}
The automatic identification, extraction and representation of the information conveyed in texts is a key task nowadays. In fact, this research topic is increasing its relevance with the exponential growth of social networks and the need to have tools that are able to automatically process them~\cite{amato2019}. 

Some of the domains where it is more important to be able to perform this kind of action are the juridical and legal ones. Effectively, it is crucial to have the capability to analyse open access text sources, like social nets (Twitter and Facebook,  for instance), blogs, online newspapers, and to be able to extract the relevant information and represent it in a knowledge base, allowing posterior inferences and reasoning.

In the context of this work, we will present results of the R\&D project Agatha\footnote{\url{http://www.agatha-osi.com/en/}}, where we  developed a pipeline of processes that analyses texts (in Portuguese, Spanish, or English) and is able to populate a specialized ontology~\cite{guarino2009} (related to criminal law) for the representation of events, depicted in such texts.  Events are represented by objects having associated actions, agents, elements, places and time. After having populated the event ontology, we have an automatic process linking the identified entities to external referents, creating, this way, a linked data knowledge base.

It is important to point out that, having the text information represented in an ontology allows us to perform complex queries and inferences, which can detect patterns of typical criminal actions.

Another axe of innovation in this research is the development, for the Portuguese language, of a pipeline of Natural Language Processing (NLP) processes, that allows us to fully process sentences and represent their content in an ontology. Although there are several tools for the processing of the Portuguese language, the combination of all these steps in a integrated tool is a new contribution.

Moreover, we have already explored other related research path, namely author profiling~\cite{Raiyani2018MultiLanguageNN}, aggression identification~\cite{W18-4404:kashyap} and hate-speech detection~\cite{semeval2019} over social media, plus statute law retrieval and entailment for Japanese~\cite{coliee2019}.

The remainder of this paper is organized as follows: Section~\ref{sec:computation_processing} describes our proposed architecture together with the Portuguese modules for its computational processing.  Section~\ref{sec:discussion} discusses different design options and Section~\ref{sec:conclusions} provides our conclusions together with some pointers for future work.

\section{Framework for Processing Portuguese Text}
\label{sec:computation_processing}

\begin{figure}[htb]
    \caption{System Overview.}\vspace{+0.2 cm}
    \centering
    \includegraphics[scale=0.35]{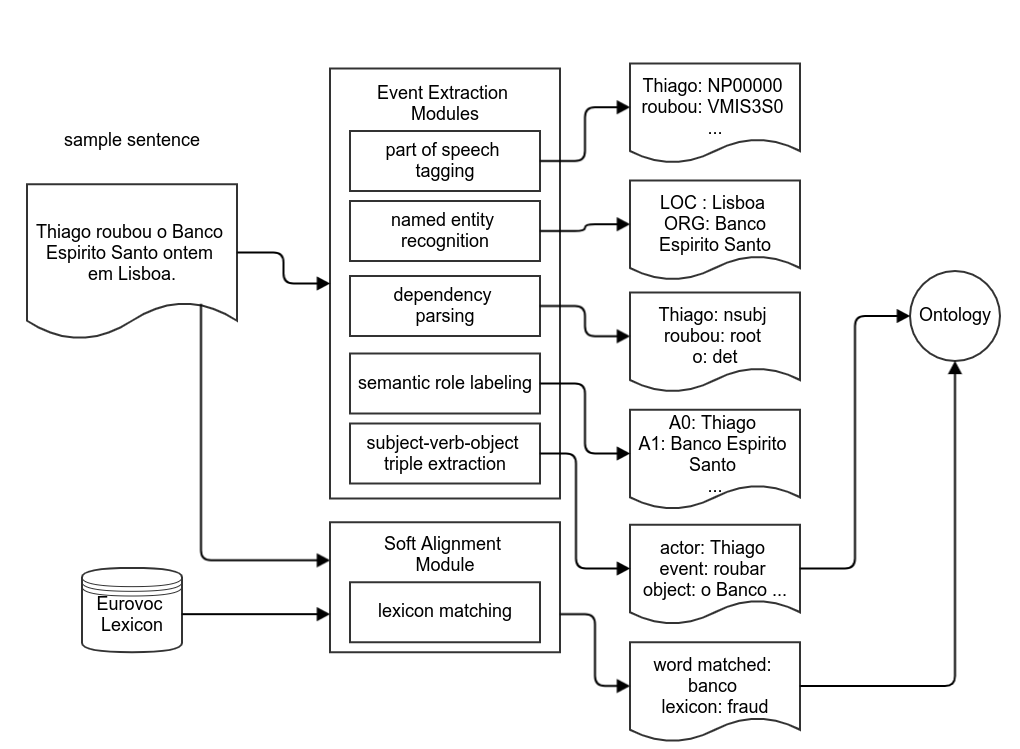}
    \label{fig:system_overview}
\end{figure}

The framework for processing Portuguese texts is depicted in Fig.~\ref{fig:system_overview}, which illustrates how relevant pieces of information are extracted from the text.  Namely, input files (Portuguese texts) go through a series of modules: part-of-speech tagging, named entity recognition, dependency parsing, semantic role labeling, subject-verb-object triple extraction, and lexicon matching. 

The main goal of all the modules except lexicon matching is to identify events given in the text. These events are then used to populate an ontology. 

The lexicon matching, on the other hand, was created to link words that are found in the text source with the data available not only on Eurovoc~\cite{eurovoc} thesaurus but also on the EU's terminology database IATE~\cite{IATE} (see Section~\ref{subsec:ontology} for details).

Most of these modules are deeply related and are detailed in the subsequent subsections.

\subsection{Part-Of-Speech Tagging}
Part-of-speech tagging happens after language detection. It labels each word with a tag that indicates its syntactic role in the sentence. For instance, a word could be a noun, verb, adjective or adverb (or other syntactic tag). We used Freeling~\cite{Lluis2004} library to provide the tags. This library resorts to a Hidden Markov Model as described by Brants~\cite{brants2000tnt}.  The end result is a tag for each word as described by the EAGLES tagset~\footnote{\url{https://talp-upc.gitbook.io/freeling-4-0-user-manual/tagsets/tagset-pt}}.

\subsection{Named Entity Recognition}
We use the named entity recognition module after part-of-speech tagging. This module labels each part of the sentence into different categories such as "PERSON", "LOCATION", or "ORGANIZATION". We also used Freeling to label the named entities and the details of the algorithm are shown in the paper by Carreras {\it et al}~\cite{carreras2003simple}.  Aside from the three aforementioned categories, we also extracted "DATE/TIME" and "CURRENCY" values by looking at the part-of-speech tags:  date/time words have a tag of "W", while currencies have "Zm".

\subsection{Dependency Parsing}
\label{subsec:dep_par}
Dependency parsing involves tagging a word based on different features to indicate if it is dependent on another word. The Freeling library also has dependency parsing models for Portuguese. Since we wanted to build a SRL (Semantic Role Labeling) module on top of the dependency parser and the current released version of Freeling does not have an SRL module for Portuguese, we trained a different Portuguese dependency parsing model that was compatible (in terms of used tags) with the available annotated.

We used the dataset from System-T~\cite{system-T}, which has SRL tags, as well as, the other preceding tags. It was necessary to do some pre-processing and tag mapping in order to make it viable to train a Portuguese model.

We made 589 tag conversions over 14 different categories. The breakdown of tag conversions per category is given by table~\ref{tab:training_development}. These rules can be  further seen in the corresponding Github repository~\cite{Universal_to_eagle_tagset}. The modified training and development datasets are also available on another Github \replaced{repository}{page}~\cite{training_and_development_dataset} for further research and comparison purposes.

\begin{table}[H]
\caption{Training and Development - Tag Set Details.}
\label{tab:training_development}
\centering
\begin{tabular}{lc}
\toprule
\textbf{Category}    & \textbf{Number of Tags} \\
\midrule
NOUN        & 20\\
VERB            & 101\\
PROPN             & 39\\
PRON             & 121\\
ADJ	        & 70\\
DET	        & 62\\
AUX        & 149\\
ADP            & 3\\
NUM             & 1\\
PUNCT	        & 18\\
CCONJ	        & 1\\
SCONJ             & 1\\
INTJ	        & 1\\
ADV	        & 2\\
\bottomrule
\end{tabular}
\end{table}

\subsection{Semantic Role Labeling}
\label{sebsec:srl}
We execute the SRL (Semantic Role Labeling) module after obtaining the word dependencies. This module aims at giving a semantic role to a syntactic constituent of a sentence. The semantic role is always in relation to a verb and these roles could either be an actor, object, time, or location, which are then tagged as A0, A1, AM-TMP, AM-LOC, respectively. We trained a model for this module on top of the dependency parser described in the previous subsection using the modified dataset from System-T.  The module also needs co-reference resolution to work and, to achieve this, we adapted the Spanish co-reference modules for Portuguese, changing the words that are equivalent (in total, we changed $253$ words).

\subsection{SVO Extraction}

From the yield of the SRL (Semantic Role Labeling) module, our framework can distinguish actors, actions, places, time and objects from the sentences. Utilizing this extracted data, we can distinguish subject-verb-object (SVO) triples using the SVO extraction algorithm~\cite{text2story}.  The algorithm finds, for each sentence, the verb and the tuples related to that verb using Semantic Role Labeling (subsection \ref{sebsec:srl}). After the extraction of SVOs from texts, they are inserted into a specific event ontology (see section \ref{subsec:ontology} for the creation of a knowledge base).

\subsection{Lexicon Matching}
\label{sebsec:lexicon}

The sole purpose of this module is to find important terms and/or concepts from the extracted text. To do this, we use Euvovoc~\cite{eurovoc}, a multilingual thesaurus that was developed for and by the European Union. The Euvovoc has 21 fields and each field is further divided into a variable number of micro-thesauri. Here, due to the application of this work in the Agatha project (mentioned in Section~\ref{sec:introduction}), we use the terms of the criminal law~\cite{1216} micro-thesaurus. Further, we classified each term of the criminal law micro-thesaurus into four categories namely, actor, event, place and object. The term classification can be seen in Table~\ref{tab:criminal_law}.

\begin{table}[H]
\caption{Eurovoc Criminal Law - Term Classification.}
\label{tab:criminal_law}
\centering
\begin{tabular}{l | c}
\toprule
\textbf{Classification}    & \textbf{$\#$ Terms} \\
\midrule
Actor & 9\\
Event & 133\\
Place & 22\\
Object & 3\\
\bottomrule
\end{tabular}
\end{table}

After the classification of these terms, we implemented two different matching algorithms between the extracted words and the criminal law micro-thesaurus terms.  The first is an exact string match wherein lowercase equivalents of the words of the input sentences are matched exactly with lower case equivalents of the predefined terms. The second matching algorithm uses Levenshtein distance~\cite{levenshtein}, allowing some near-matches that are close enough to the target term.

\subsection{Linked Data: Ontology, Thesaurus and Terminology}
\label{subsec:ontology}

In the computer science field, an ontology can be defined has:
\begin{itemize}
    \item a formal specification of a conceptualization;
    \item shared vocabulary and taxonomy which models a domain with the definition of objects and/or concepts and their properties and relations;
    \item the representation of entities, ideas, and events, along with their properties and relations, according to a system of categories.
\end{itemize}    

A knowledge base is one kind of repository typically used to store answers to questions or solutions to problems enabling rapid search, retrieval, and reuse, either by an ontology or directly by those requesting support.  For a more detailed description of ontologies and knowledge bases, see for instance~\cite{Uniti1995}.

\begin{figure}
    \centering
    \includegraphics[scale=0.5]{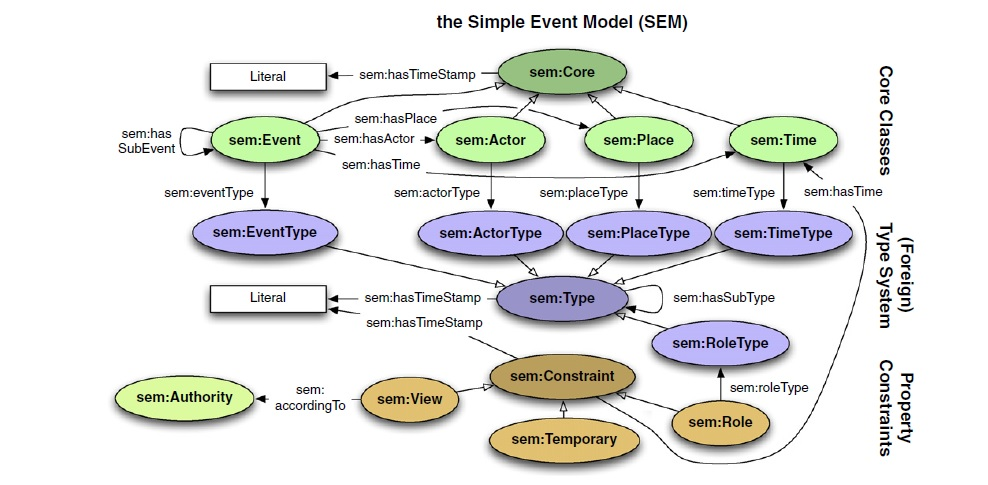}
    \caption{The Simple Event Model~\cite{van2011design}}
    \label{fig:sem}
\end{figure}

For designing the ontology adequate for our goals, we referred to the Simple Event Model (SEM)~\cite{van2011design} as a baseline model. A pictorial representation of this ontology is given in Figure~\ref{fig:sem}

Considering the criminal law domain case study, we made a few changes to the original  SEM ontology. The entities of the model are:
\begin{itemize}
    \item Actor: person involved with event
    \item Place: location of the event
    \item Time: time of the event
    \item Object: that actor act upon
    \item Organization: organization involved with event
    \item Currency: money involved with event
\end{itemize}

The proposed ontology was designed in such a manner that it can incorporate information extracted from multiple documents.
\added{In this context, suppose that} the source of documents \replaced{is a}{are a legal} police department, where each document \replaced{is}{are} under the hood of a particular case/crime; \replaced{furthermore}{Further}, a single case can have documents from multiple languages. Now, considering case 1 has 100 documents and case 2 has 100 documents then there is not only a connection among the documents of a single case but rather among all the cases with all the combined 200 documents. In this way, the proposed method is able to produce a detailed and well-connected knowledge base. 

\begin{figure}[htbp]
    \caption{Ontology Diagram.}\vspace{+0.2 cm}
    \centering
    \includegraphics[width=1.0\textwidth]{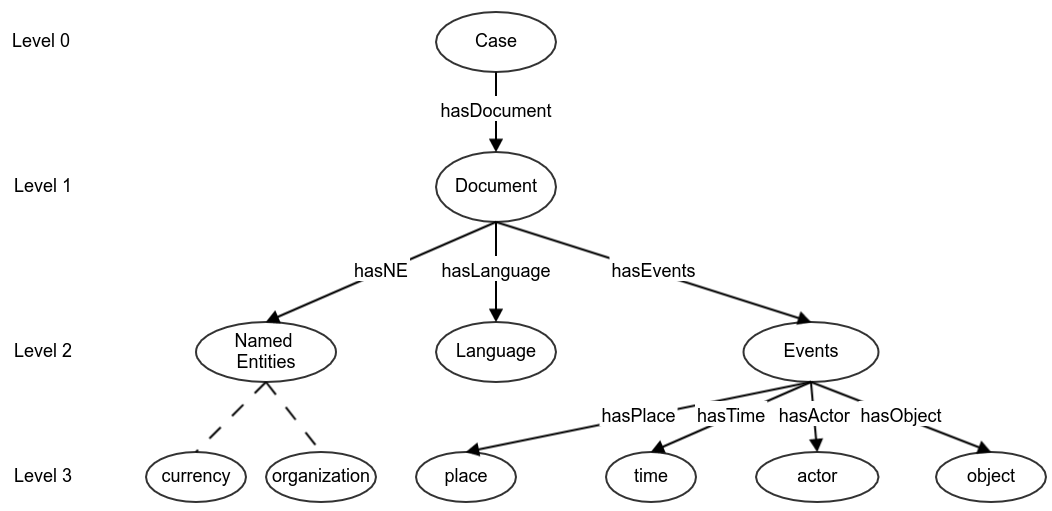}
    \label{fig:ontology}
\end{figure}

Figure~\ref{fig:ontology} shows the proposed ontology, which, in our evaluation procedure, was populated with 3121 events entries from 51 documents.

Protege~\cite{protege} tool was used for creating the ontology and GraphDB~\cite{graphDB} for populating \& querying the data. GraphDB is an enterprise-ready Semantic Graph Database, compliant with W3C Standards. Semantic Graph Databases (also called RDF triplestores) provide the core infrastructure for solutions where modeling agility, data integration, relationship exploration, and cross-enterprise data publishing and consumption are important. GraphDB has a SPARQL (SQL-like query language) interface for RDF graph databases with the following types:
\begin{itemize}
    \item SELECT: returns tabular results
    \item CONSTRUCT: creates a new RDF graph based on query results
    \item ASK: returns "YES", if the query has a solution, otherwise "NO"
    \item DESCRIBE: returns RDF data about a resource. This is useful when the RDF data structure in the data source is not known
    \item INSERT: inserts triples into a graph
    \item DELETE: deletes triples from a graph
\end{itemize}

Furthermore, we have extended the ontology~\cite{extended_ontology} to connect the extracted terms with Eurovoc criminal law (discussed in subsection \ref{sebsec:lexicon}) and IATE~\cite{IATE} terms. IATE (\textbf{I}nter\textbf{a}ctive \textbf{T}erminology for \textbf{E}urope) is the EU's general terminology database and its aim  is to provide a web-based infrastructure for all EU terminology resources, enhancing the availability and standardization of the information. The extended ontology has a number of sub-classes for Actor, Event, Object and Place classes detailed in  Table~\ref{tab:ontology_subclass}.

\begin{table}[htbp]
\caption{Extended Ontology~\cite{extended_ontology} - Sub-Classes.}
\label{tab:ontology_subclass}
\centering
\begin{tabular}{ccc}
\toprule
\textbf{Class}  & \textbf{\makecell{No. of\\ Sub-Classes}} & \textbf{Terms} \\
\midrule
Actor & 6 & Victim,Inmate,Prisoner,Hostage,Hijacker,Accomplice\\
\hline
Event & 64 & \makecell{Slavery,Trade,Tax,Evasion,Spoofing,Slander,Shady,\\Violence,Sexual,Scam,Repentance,Rehabilitation,\\Refoulement,Rape,Punishment,Ponzi,Piracy,Aggression,\\Phishing,Trafficking,Pardon,Harassment,Mobbing,\\Misdemeanour,Libel,Trading,Imprisonment,\\Restraint,Theft,Arrest,Homicide,Hit-and-run,Hijacking,\\Forgery,Forfeiture,Fraud,Fight,Falsification,Extradition,\\Expulsion,Elimination,Offence,Drug,Detention,\\Deprivation,Deportation,Defamation,Penalty,Negligence,\\Execution,Counterfeit,Corruption,Confiscation,\\Conditional,Con,Order,Complicity,Campaign,\\Bully,Breach,Banish,Aggravate,Crime,Abduction} \\
\hline
Object & 3 & Fine,Invoice,Bill \\
\hline
Place & 11 & \makecell{Facility,Institution,Center,Confinement,Reformatory\\Penitentiary,Penal,Prison,Jail,Isolation,Banco}\\
\bottomrule
\end{tabular}
\end{table}

\section{\added{Discussion}}
\label{sec:discussion}

We have defined a major design principle for our architecture: it should be modular and not rely on human made rules allowing, as much as possible, its independence from a specific language.
In this way, its potential application to another language would be easier, simply by 
changing the modules or the models of specific modules.
In fact, we have explored the use of already existing modules and adopted and integrated several of these tools into our pipeline.

It is important to point out that, as far as we know, there is no integrated
architecture supporting the full processing pipeline for the Portuguese language. 
We evaluated several systems like Rembrandt~\cite{cardoso-2012-rembrandt} or LinguaKit: the former only has the initial steps of our proposal (until NER) and the later performed worse than our system.

This framework, developed within the context of the Agatha project (described in Section~\ref{sec:introduction}) has the full processing pipeline for Portuguese
texts: it receives sentences as input and outputs ontological information:
a) first performs all NLP typical tasks until semantic role labelling; 
b) then, it extracts subject-verb-object triples; 
c) and, then, it performs
ontology matching procedures. As a final result, the obtained output is inserted into
a specialized ontology.

We are aware that each of the architecture modules can, and should, be improved
but our main goal was the creation of a full working text processing pipeline for
the Portuguese language.

\section{Conclusions and Future Work}
\label{sec:conclusions}

Besides the end--to--end NLP pipeline for the Portuguese language, the other main contributions of this work can be summarize as follows:
\begin{itemize}
    \item Development of an ontology for the criminal law domain;
    \item Alignment of the Eurovoc thesaurus and IATE terminology with the ontology created;
    \item Representation of the extracted events from texts  in the linked knowledge base defined. 
\end{itemize}

The obtained results support our claim that the proposed system can be used as
a base tool for information extraction for the Portuguese language.
Being composed by several modules, each of them with a high level of complexity,
it is certain that our approach can be improved and an overall better performance
can be achieved.

As future work we intend, not only to continue improving the individual modules, but also plan to extend this work to the:
\begin{itemize}
    \item automatic creation of event timelines;
    \item incorporation in the knowledge base of information obtained from videos or pictures describing scenes relevant to criminal investigations.
\end{itemize}

\section*{Acknowledgments}
The authors would like to thank COMPETE 2020, PORTUGAL 2020 Program, the European Union, and ALENTEJO 2020 for supporting this research as part of Agatha Project SI \& IDT number 18022 (Intelligent analysis system of open of sources information for surveillance/crime control).  The authors would also like to thank LISP - Laboratory of Informatics, Systems and Parallelism.
%
%
\bibliographystyle{splncs04}
\bibliography{mybibliography}

\begin{thebibliography}{10}
\providecommand{\url}[1]{\texttt{#1}}
\providecommand{\urlprefix}{URL }
\providecommand{\doi}[1]{https://doi.org/#1}

\bibitem{Universal_to_eagle_tagset}
Automated event extraction model for multiple linked portuguese documents,
  \url{https://github.com/kraiyani/Automated-Event-Extraction-Model-for-Multiple-Linked-Portuguese-Documents/blob/master/Universal_to_eagle_tagset.xlsx},
  [Available online: accessed on 06 05 2019]

\bibitem{eurovoc}
Eu vocabularies, \url{https://publications.europa.eu/en/web/eu-vocabularies},
  [Available online: accessed on 06 05 2019]

\bibitem{1216}
Eu vocabularies, thesauri, 1216 criminal law,
  \url{https://publications.europa.eu/en/web/eu-vocabularies/th-concept-scheme/-/resource/eurovoc/100180?target=Browse},
  [Available online: accessed on 06 05 2019]

\bibitem{extended_ontology}
Extended ontology, \url{http://owlgred.lumii.lv/online_visualization/e9fh#},
  [Available online: accessed on 25 06 2019]

\bibitem{graphDB}
Graphdb, \url{http://graphdb.ontotext.com/}, [Available online: accessed on 06
  05 2019]

\bibitem{IATE}
Iate (interactive terminology for europe), \url{https://iate.europa.eu/home},
  [Available online: accessed on 06 05 2019]

\bibitem{levenshtein}
Levenshtein distance, \url{https://en.wikipedia.org/wiki/Levenshtein_distance},
  [Available online: accessed on 06 05 2019]

\bibitem{system-T}
Portuguese universal propositions,
  \url{https://github.com/System-T/UniversalPropositions/tree/master/UP_Portuguese-Bosque},
  [Available online: accessed on 06 05 2019]

\bibitem{protege}
Protege, \url{https://protege.stanford.edu/}, [Available online: accessed on 06
  05 2019]

\bibitem{training_and_development_dataset}
Training and development dataset for automated event extraction model for
  multiple linked portuguese documents,
  \url{https://github.com/kraiyani/Automated-Event-Extraction-Model-for-Multiple-Linked-Portuguese-Documents},
  [Available online: accessed on 06 05 2019]

\bibitem{amato2019}
Amato, F., Moscato, V., Picariello, A., Sperl{\`{\i}}, G.: Extreme events
  management using multimedia social networks. Future Generation Comp. Syst.
  \textbf{94},  444--452 (2019). \doi{10.1016/j.future.2018.11.035},
  \url{https://doi.org/10.1016/j.future.2018.11.035}

\bibitem{brants2000tnt}
Brants, T.: Tnt: a statistical part-of-speech tagger. In: Proceedings of the
  sixth conference on Applied natural language processing. pp. 224--231.
  Association for Computational Linguistics (2000)

\bibitem{cardoso-2012-rembrandt}
Cardoso, N.: Rembrandt - a named-entity recognition framework. In: Proceedings
  of the Eighth International Conference on Language Resources and Evaluation
  ({LREC}-2012). pp. 1240--1243. European Language Resources Association
  (ELRA), Istanbul, Turkey (May 2012),
  \url{http://www.lrec-conf.org/proceedings/lrec2012/pdf/409\_Paper.pdf}

\bibitem{Lluis2004}
Carreras, X., Chao, I., Padr{\'o}, L., Padro, M.: Freeling: An open-source
  suite of language analyzers. Proceedings of the 4th International Conference
  on Language Resources and Evaluation (LREC'04)  (01 2004)

\bibitem{carreras2003simple}
Carreras, X., M{\`a}rquez, L., Padr{\'o}, L.: A simple named entity extractor
  using adaboost. In: Proceedings of the seventh conference on Natural language
  learning at HLT-NAACL 2003 (2003)

\bibitem{Uniti1995}
Guarino, N., Giaretta, P.: Ontologies and knowledge bases: Towards a
  terminological clarification. In: Towards very Large Knowledge bases:
  Knowledge Building and Knowledge sharing. pp. 25--32. IOS Press (1995)

\bibitem{guarino2009}
Guarino, N., Oberle, D., Staab, S.: What Is an Ontology?, pp. 1--17 (05 2009)

\bibitem{W18-4404:kashyap}
Raiyani, K., Gon{\c{c}}alves, T., Quaresma, P., Nogueira, V.B.: Fully connected
  neural network with advance preprocessor to identify aggression over facebook
  and twitter. In: Proceedings of the First Workshop on Trolling, Aggression
  and Cyberbullying (TRAC-2018). pp. 28--41. Association for Computational
  Linguistics (2018), \url{http://aclweb.org/anthology/W18-4404}

\bibitem{Raiyani2018MultiLanguageNN}
Raiyani, K., Gon\c{c}alves, T., Quaresma, P., Nogueira, V.B.: Multi-language
  neural network model with advance preprocessor for gender classification over
  social media: Notebook for pan at clef 2018. In: Working Notes of {CLEF} 2018
  - Conference and Labs of the Evaluation Forum, Avignon, France, September
  10-14, 2018. (2018), \url{http://ceur-ws.org/Vol-2125/paper\_105.pdf}

\bibitem{text2story}
Raiyani, K., Gon\c{c}alves, T., Quaresma, P., Nogueira, V.B.: Automated event
  extraction model for linked portuguese documents. In: Proceedings of
  Text2Story — Second Workshop on Narrative Extraction From Texts co-located
  with 41th European Conference on Information Retrieval (ECIR 2019) Cologne,
  Germany, April 14th (2019), \url{http://ceur-ws.org/Vol-2342/paper2.pdf}

\bibitem{semeval2019}
Raiyani, K., Gon\c{c}alves, T., Quaresma, P., Nogueira, V.B.: Vista.ue at
  semeval-2019 task 5: Single multilingual hate speech detection model. In:
  Proceedings of the 13th International Workshop on Semantic Evaluation
  (SemEval-2019). pp. 520--524. Association for Computational Linguistics
  (2019)

\bibitem{coliee2019}
Raiyani, K., Quaresma, P.: Keyword \& machine learning based japanese statute
  law retrieval and entailment task at coliee-2019. In: Proceedings of
  Competition on Legal Information Retrieval and Entailment Workshop (COLIEE
  2019) in association with the 17th International Conference on Artificial
  Intelligence and Law 2019 (ICAIL 2019). Easychair (2019)

\bibitem{van2011design}
Van~Hage, W.R., Malais{\'e}, V., Segers, R., Hollink, L., Schreiber, G.: Design
  and use of the simple event model (sem). Web Semantics: Science, Services and
  Agents on the World Wide Web  \textbf{9}(2),  128--136 (2011)

\end{thebibliography}

\end{document}